\documentclass[letterpaper, 10 pt, journal, twoside]{IEEEtran}

\IEEEoverridecommandlockouts

\usepackage{amsmath}
\usepackage{float}
\usepackage{graphicx}
\usepackage{algorithm, algpseudocode}
\usepackage{footnote}
\usepackage{lipsum}
\usepackage{enumitem}
\usepackage{comment}
\usepackage{xfrac}
\usepackage{multirow}
\usepackage{bm}
\usepackage{color}
\usepackage[normal]{subfigure}
\usepackage{cite}

\title{Learning Densities in Feature Space for Reliable Segmentation of Indoor Scenes}

\author{Nicolas Marchal$^{*}$, Charlotte Moraldo$^{*}$, Hermann Blum, Roland Siegwart, Cesar Cadena, Abel Gawel \thanks{* Authors contributed equally.}
\thanks{Manuscript received: September, 10, 2019; Revised December, 17, 2019; Accepted January, 9, 2020.}
\thanks{This paper was recommended for publication by Editor Eric Marchand upon evaluation of the Associate Editor and Reviewers' comments.
This work was partially supported by the HILTI group.}
\thanks{Authors are with the Autonomous Systems Lab, ETH Zurich. {\tt\footnotesize \{marchaln, moraldoc, rsiegwart, blumh, cesarc, gawela\}@ethz.ch}}
}

\begin{document}

\maketitle

\begin{abstract}
Deep learning has enabled remarkable advances in scene understanding, particularly in semantic segmentation tasks. Yet, current state of the art approaches are limited to a closed set of classes, and fail when facing novel elements, also known as out of distribution (OoD) data. This is a problem as autonomous agents will inevitably come across a wide range of objects, all of which cannot be included during training. We propose a novel method to distinguish any object (\textit{foreground}) from empty building structure (\textit{background}) in indoor environments. We use normalizing flow to estimate the probability distribution of high-dimensional background descriptors. Foreground objects are therefore detected as areas in an image for which the descriptors are unlikely given the background distribution. As our method does not explicitly learn the representation of individual objects, its performance generalizes well outside of the training examples. Our model results in an innovative solution to reliably segment foreground from background in indoor scenes, which opens the way to a safer deployment of robots in human environments.
\end{abstract}


\begin{IEEEkeywords}
Deep Learning in Robotics and Automation, Semantic Scene Understanding, Visual Learning
\end{IEEEkeywords}


\section{Introduction}

\IEEEPARstart{D}{eep} learning methods have allowed significant improvements in computer vision and semantic segmentation tasks in robotic applications \cite{denseobjectnets}. Yet, an important drawback of current Deep Neural Networks trained for classification or segmentation is that they are trained to recognize a fixed set of classes with a limited number of examples. Thus, they behave poorly and unpredictably when they are given out-of-distribution (OoD) examples \cite{metrics, ood2, ood3}. Moreover, in such cases, networks often give wrong predictions with high confidence. Although a majority of visual recognition systems are designed for a static closed world, these algorithms aim to be deployed in the dynamic and ever-changing real world \cite{openset}. The diversity and variability of our world makes algorithms designed to perform static closed set recognition unsafe and limits the deployment of robotic systems in our everyday tasks \cite{safety}. This is particularly important in indoor environments, where diverse objects are often added, moved, or altered. Most autonomous robots designed to operate in the presence of humans will be in such settings. In indoor scenes, background consists of the basic room structure (floor, walls, ceiling, windows), which remains static. Foreground then contains all the dynamic elements (objects, furniture, people etc.), which are prone to much more variability and novelty. State-of-the-art systems rely on closed-set classification to find potentially dynamic objects \cite{Xu2018-tn}. For reliable applications of indoor robotics, it is critical to segment such dynamic or novel elements from the background rather than identifying their nature. Classical semantic segmentation is not well suited for this task as it relies on the strong assumption that all objects in the foreground are known a priori \cite{openset}. A good way of separating all dynamic objects from the static background would be to use a reliable binary background-foreground segmentation.

\begin{figure}
\centering
\includegraphics[width=0.9\columnwidth]{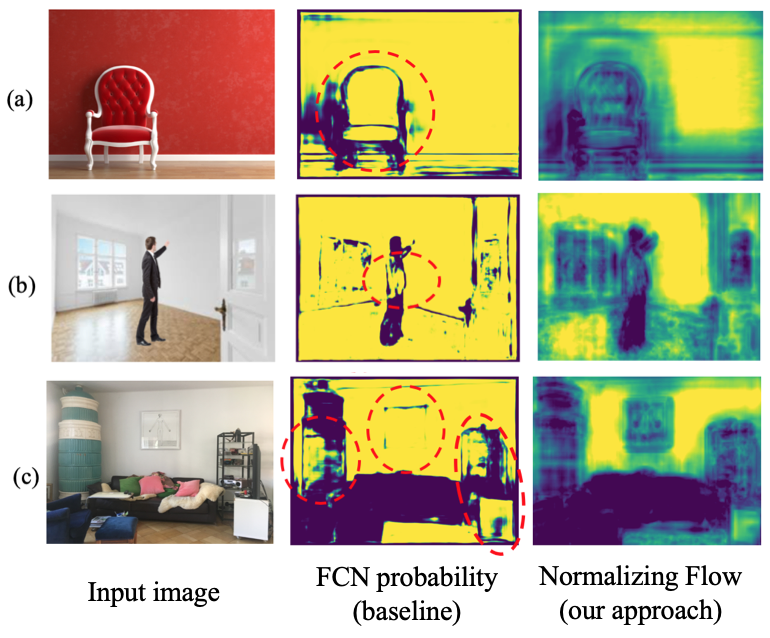}
\vspace{-1mm}
\caption{Example results of the proposed feature density-based segmentation method demonstrating its robustness against out-of-distribution (OoD) data (red dotted circles were added to highlight failures of the baseline FCN softmax method): (a) Advertisement image where part of the chair is missed by the FCN softmax. (b) Picture from Google Image where we add a human (people are not part of our training set). (c) Smartphone picture of a living room: several elements are missed by the FCN softmax.} 
\label{fig:method_test}
\vspace{-0.2cm}
\end{figure}

We introduce a novel method to segment foreground from background, without limiting foreground objects to a fixed set. We extract feature descriptors of the background in images with a convolutional neural network (CNN). The descriptors, which are the outputs of some deep layer in the CNN, characterise the background appearance in a high dimensional space with a complex unknown distribution. Leveraging recent advances in flow-based generative models \cite{NICE, NVP, GLOW}, we use normalizing flow to model this distribution, as done in \cite{PE}. Foreground objects are then recognized as having descriptors unlikely to come from the estimated background distribution. The novelty of our method is that the training of the segmentation relies solely on background, which has a low variability. Its performance is therefore less affected by the high variability of foreground. Moreover our method can segment an unlimited set of objects, as long as their appearance differ from the background. To assess the reliability of our method, we create a dataset containing novel objects, which we coined \textit{generalization dataset}. The performance drop between the test and generalization datasets on all evaluated metrics is 1.5 to 10 times larger for a classic FCN softmax segmentation compared to our method. 

We benchmark our results by comparing them with a k-Nearest-Neighbours (kNN) kernel density approach and the softmax score of a classical Fully Convolutional Network (FCN) segmentation. Using normalizing flows proved to i) be a completely novel approach to binary semantic segmentation, ii) be more reliable than a classical encoder decoder network with softmax segmentation, iii) be computationally more efficient than kNN and iv) yield better estimates of high dimensional density distributions than kNN. 

\section{Related Work}
\label{sec:rel-work}

\subsection{Supervised Semantic Segmentation and Object Detection} 
Leveraging recent advances in deep learning, state-of-the-art methods in semantic segmentation are comprised of fully-convolutional networks trained with pixelwise supervision \cite{fcn8}. These methods use an encoder-decoder architecture \cite{decoder}, where the role of the decoder network is to map the low resolution encoder features to full input resolution pixelwise classification. In our work, we estimate the probability that the encoder's features were generated by background to create a background likelihood map, which we can then segment to obtain a binary pixelwise semantic segmentation.

Semantic segmentation techniques are commonly used for object detection. For example, Mask R-CNN \cite{maskrcnn} uses faster R-CNN \cite{object4} to generate Regions of Interest (ROIs) likely to contain objects and then predicts per-pixel semantically annotated masks inside all ROIs. \cite{reconstruction2} uses depth information to refine the predictions of Mask R-CNN and build 3D maps of indoor environments. Our method does not rely on any object detector like faster R-CNN and does not use depth information. \cite{segnet} remap the 894 labels of NYU into 14 classes (objects,  furniture,  wall, ceiling etc.) to perform semantic segmentation. This illustrates one of the big drawbacks of classical object detectors: they are limited to a fixed set of classes. In contrast, our method is designed to work with an unlimited number of different foreground objects. Another drawback of classical segmentation methods is that they need a large amount of labeled data, which is expensive and difficult to obtain. Large amounts of labeled data are required for segmentation to capture the variability and complexity in foreground objects. However, since our approach relies solely on background, we need fewer data and we can use images of fully empty rooms which do not need to be labeled.  

While a lot of research aims to improve the efficiency of object detection \cite{object4} and semantic segmentation, few of them focus on their reliability to OoD data. Moreover, despite the amount of work in object detection and scene understanding, the related problem of detecting all foreground objects has not yet been fully addressed. Our results show that our solution provides a way to increase reliability of binary segmentation networks, allowing better generalization on OoD data.

 Background modeling methods have been used in videos, for background removal \cite{back_video2} or object localization \cite{back_video3}. These methods however make use of the movement in videos and are not applicable for object segmentation in single images.

\subsection{Novelty and Out-of-Distribution (OoD) Detection}
Although Bayesian deep learning allows uncertainty representation in settings such as regression or classification \cite{bayesian2, bayesian3}, non-Bayesian approaches for novelty detection have recently become more popular. \cite{deepkNN} for instance combines the k-nearest neighbors algorithm with representations of the data learned by each layer of a neural network to identify Ood data. Computationally more efficient alternatives include density estimation and generative probabilistic modelling methods \cite{ood3, NVP, GLOW}, which allow one to estimate the likelihood of samples with respect to the true data distribution and learn meaningful features while requiring little supervision or labeling. \cite{PE} shows great potential in using flow-based approaches \cite{NVP} to approximate the probability distribution of deep convolutional features to identify OoD data inside images, generating a binary segmentation between known and unknown data. Inspired by these works, we propose a novel application of normalizing flow for reliable segmentation of foreground and background. 

\subsection{Normalizing Flow}
\label{subsec:prior_normalizing_flow}
The data of interest in deep learning frameworks is generally high-dimensional and highly structured. The challenge in modelling high-dimensional densities for this data is that we need models powerful enough to capture its complexity but yet still be trainable and efficient. Flow-based models, first described in \cite{NICE} and later improved in \cite{NVP} and \cite{GLOW} are known for their generative properties but can also be used for efficient and exact computation of high dimensional density functions, using the change of variable formula. We use this tool to estimate the probability distribution of background features.

\section{Method}
\label{sec:method}

Our work introduces a new way of segmenting background from foreground in indoor scenes. Prior approaches use labeled datasets to learn pixelwise classifications given foreground objects and background scenes in a training set. Although these approaches work well when scenes contain the objects they were trained for, their behavior is unpredictable if we introduce new elements. To tackle this issue, we suggest a different method that learns only the background appearance of indoor rooms. This is done in two steps:

\begin{itemize}[noitemsep, leftmargin=*]
\item We use a CNN, coined \emph{expert network}, to generate features from images (Section \ref{sec:feature_extraction}). Our descriptors (or features) are simply the output of a convolution layer in the expert network, and are thus points in a high-dimensional space.

\item We then learn what the background in our training set ``looks like''. To do so, we use normalizing flow \cite{NVP} to estimate the probability distribution of the background features extracted by the expert network (Section \ref{sec:density}).
\end{itemize} 

When feeding an image of dimension $H \times W \times C$ ($H$ the height, $W$ the width and $C$ the number of channels) to our network, the expert model transforms it to a feature map of size $h \times w \times f$ (where, for 2x2 pooling filters, $h = $ \sfrac{H}{\text{2 (\# pooling layers)}},  $w = $ \sfrac{W}{\text{2 (\# pooling layers)}} and $f$ is the feature dimension, which depends on the architecture of the expert model). Using our approximated background feature distribution, we can transform the feature map into a likelihood map of dimension $h \times w \times 1$. We furthermore use bilinear interpolation to upscale the likelihood map back to the original image dimensions ($H \times W \times 1$). Foreground objects are detected as areas of low probability in the image and a binary segmentation is obtained by thresholding the density map. 

\begin{figure}[t]
\centering
\subfigure[Feature extraction: generate background features from indoor images by extracting the output of a convolution layer in the expert network. The 6\textsuperscript{th} layer is a combination of the 4\textsuperscript{th} and the (upsampled) 5\textsuperscript{th} layer.]{\includegraphics[width=0.96\columnwidth]{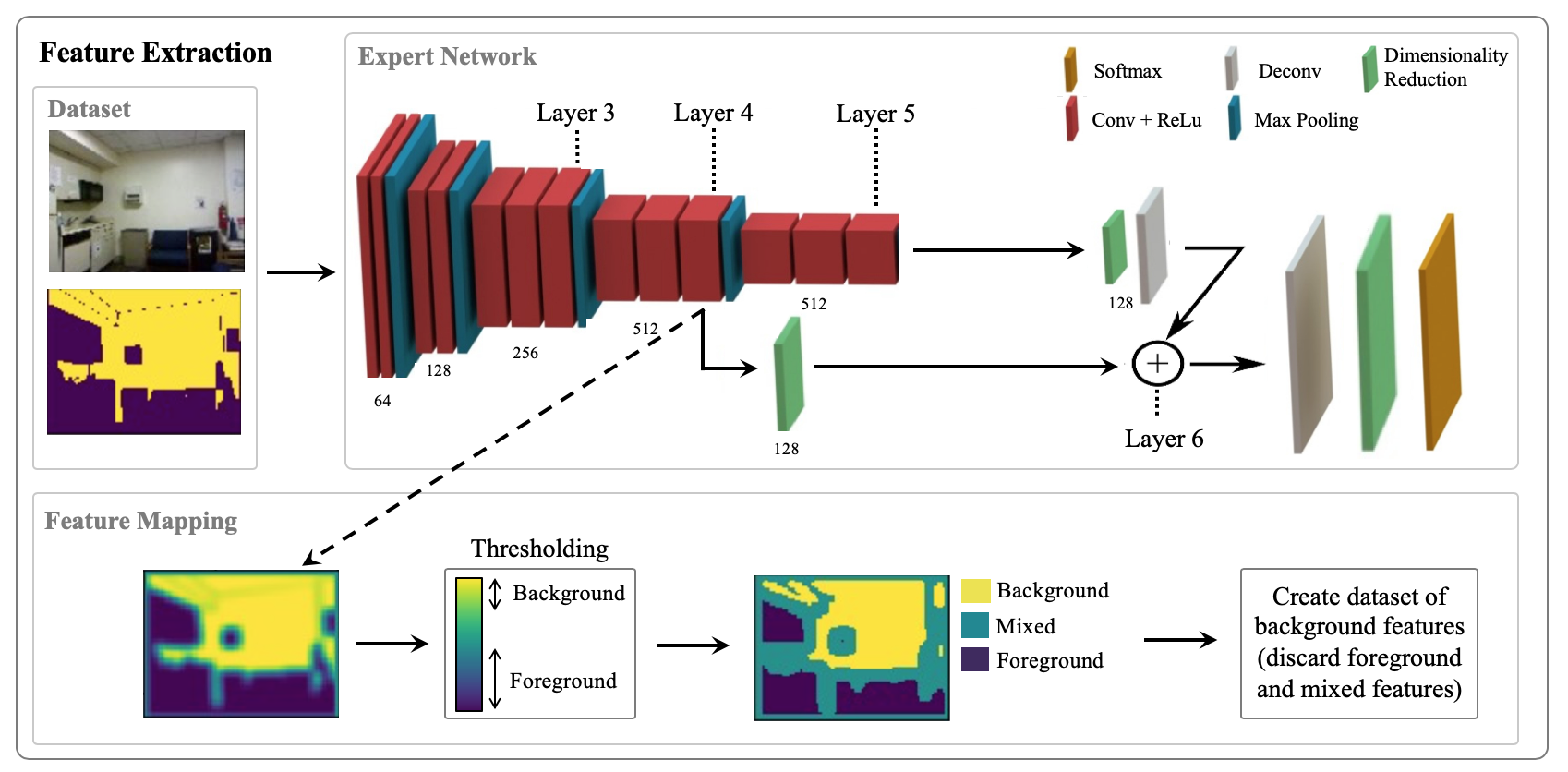}
\label{fig:method_a}}
\hfill
\subfigure[Density estimation: estimate the probability distribution of the background features using normalizing flow]{\includegraphics[width=0.96\columnwidth]{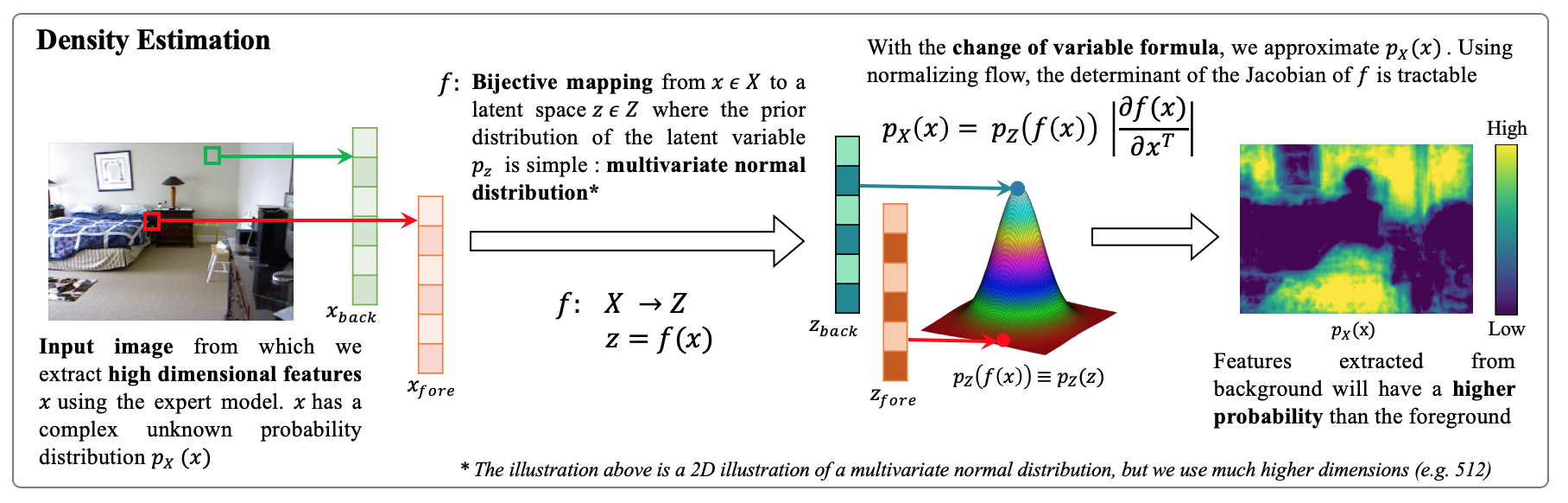}
\label{fig:method_b}}
\hfill
\caption{Illustration of our method where (a) shows how to extract features and (b) shows how to estimate the background likelihood from features.} 
\label{fig:method}
\end{figure}

The strength of our method is that it detects foreground objects without ever explicitly learning their appearance. It therefore does not require a training set of foreground elements and does not limit us to recognizing a fixed set of objects. Our complete pipeline is illustrated in Fig. \ref{fig:method}.

\subsection{Feature extraction}
\label{sec:feature_extraction}
\subsubsection{Collecting background data}
Ideally, we would train our network on fully empty rooms, but such a dataset does not exist. Instead, we use a dataset of labeled indoor images (NYU Depth v2 \cite{NYU}) and generate binary masks to differentiate the background from the foreground. The dataset contains a large number of specific labels, which we map to background or foreground. We define background to be what an empty room would contain (ceiling, floor, wall, window). Any other object is then considered as foreground. With this label mapping, we create a binary mask for each image of the dataset.

\subsubsection{Expert Network}
Our expert network is a fully-convolutional network (FCN) \cite{fcn8}, which consists of a VGG-16 encoder \cite{vgg16} followed by a deconvolution. This is a classical architecture for segmentation tasks, which we use in three ways: (i) given the NYU dataset with binary masks as described above, we train this network to perform background-foreground segmentation; (ii) we also show that training of the expert model is not necessary by using weights from a standard VGG-16 trained on ImageNet (note that we use the weights from ImageNet for the encoder but we still train the decoder); iii) similarly, we also use weights from an encoder trained for scene parsing on the ADE20K dataset \cite{ADE20K}. In this work, we refer to these three approaches as using (i) NYU segmentation, (ii) ImageNet and (iii) ADE20K. Note that these encoders have been trained for different tasks and with different datasets. We discuss their results in sections \ref{subsec:results_test} and \ref{subsec:generalization}. 

\subsubsection{Feature Mapping}
We extract features from a chosen layer of the expert network. After convolutions and pooling, features from deep layers will have a large receptive field and are thus likely to have been affected by both background and foreground. To determine which label has most influenced the features, we give them a score between 0 and 1: a high score represents a large proportion of background pixels in the receptive field. We then use a threshold to assign the labels ``foreground'', ``background'' or ``mixed'' to all features, as shown on Fig. \ref{fig:method_a}. Finally, we discard features with foreground and mixed labels to create a dataset of background features, on which we perform density estimation.

\subsection{Density Estimation}
\label{sec:density}
Given a dataset of background features $\bm{X}$, we estimate the probability $p_X(x)$ of given image features $x$ to belong to the background class. To estimate the probability density, we compare kNN and normalizing flow. From the probability for each feature vector, that encodes a patch in the input image, we linearly interpolate the probability to pixel-resolution.

\subsubsection{kNN Density Estimation} 
Different works \cite{deepkNN, distancebased} use the distance between a test feature $x$ and its $k$ nearest training features to estimate uncertainty in neural networks. They show that distance in feature space is an informative metric to identify OoD data. Similarly, we use kNN kernel density to approximate $p_X(x)$. We compute a likelihood using the distances to the $k$ nearest background representations:
\begin{equation*}
    p_X(x) \propto \frac{1}{k} \sum_{k}^{} \exp \big( -dist_k(x, \bm{X} \big)\, )
\end{equation*}
Following ~\cite{deepkNN, PE}, we use the cosine similarity as distance metric. Intuitively, if a feature lies in a part of the space far from any background feature, it is very unlikely to correspond to background. Unfortunately, this method requires kNN lookups for every image patch, which is computationally inefficient.

\subsubsection{Flow-Based Density Estimation}
Flow-based approaches have proven useful for estimating complex high dimensional densities~\cite{NVP}. The method aims to find a bijective transformation $f$ that maps feature vectors $x$ to a latent space $z$, where $p_Z$ is the prior probability distribution of the latent variable. $p_Z$ is chosen to be a simple and tractable density such as a multivariate Gaussian. $f$ is constructed as a chain of bijective transformations: $f = f_1 \circ f_2 \circ ... \circ f_n$ which is called a (normalizing) flow. As explained in \cite{NVP}, the scale and translation operations in the bijective transformations $f_i$ are arbitrarily complex functions, thus modeled as deep neural networks with a set of weights $\theta$. We call $p_{\theta}$ the approximation of $p_X$. $\theta$ is learned by minimizing the negative log likelihood (NLL) of all background features. 
\begin{equation}
\begin{split}
\begin{aligned}
    &\min_{\theta} \enspace  - \frac{1}{\bm{|X|}} \sum_i \log p_{\theta}(x^{(i)}) \qquad \quad \text{where:} \\
    &\log (p_{\theta}(x)) = \log \big(p_Z(f(x)) \big) + \log \Big( \Big| \frac{\partial f(x)}{\partial x^T} \Big| \Big)
\end{aligned}
\end{split}
\label{eq:flow}
\end{equation}

The NLL is obtained by using the change of variable formula (eq. \ref{eq:flow}), and requires efficient computation of the determinant of $\frac{\partial f(x)}{\partial x^T}$ (Jacobian of $f$). NICE \cite{NICE} introduces a family of bijective transformations called coupling layers for which the Jacobian is a triangular matrix: its determinant is therefore tractable and can be efficiently computed. Real NVP \cite{NVP} and GLOW \cite{GLOW} further improve the work of NICE \cite{NICE}, allowing efficient and exact log density estimation of data points $x$. In this work, we use Real NVP \cite{NVP} with a chain length of 32 to obtain an efficient density estimation of our background features.


The benefits of normalizing flow over kNN are: (i) it can learn more complex distributions, and (ii) the memory footprint of the flow weights is lower than a kNN database. 


\subsubsection{Multiple Feature Layers}
To perform density estimation at multiple layers of our expert network's encoder, we need to combine the likelihoods from an ensemble of density estimators, similar to \cite{ood2}. However, the individual negative log-likelihood (NLL) estimates cannot be aggregated because the different background features distributions have varying dispersion and dimensions. Similar problems arise for the cosine distance of kNN density. Densities at different layers thus have different scales. Similar to \cite{PE}, we first center the NLL at layer $l$ around the average NLL of the training features  for that layer: $\Bar{N}(z_l) = N(z_l) - \mathcal{L} (Z_l)$. In the ideal case of a multivariate Gaussian, $\Bar{N}$ corresponds  to  the  Mahalanobis distance used in \cite{ood2}. Using a small validation set, we estimate the mean and standard deviation to normalize $\Bar{N}$ such that the NLL of individual layers can be compared. Following \cite{PE}, we experiment with three strategies: (i) A pixel is detected as foreground only if all layers agree that it has low log likelihood (high NLL), thus having a high minimum NLL. (ii) A pixel is detected as foreground if it has a low log likelihood on at least one layer, thus having a high maximum NLL. (iii) Using a small set of new labeled images, we fit a logistic regression to capture the interaction between the layers and use it to improve the individual predictions. The results and discussion will be centered around iii) since it outperformed significantly i) and ii) in our experiments.

\section{Experiments and Results}
\label{sec:result}

\begin{figure*}[t]
	\centering
    \includegraphics[width=1.9\columnwidth]{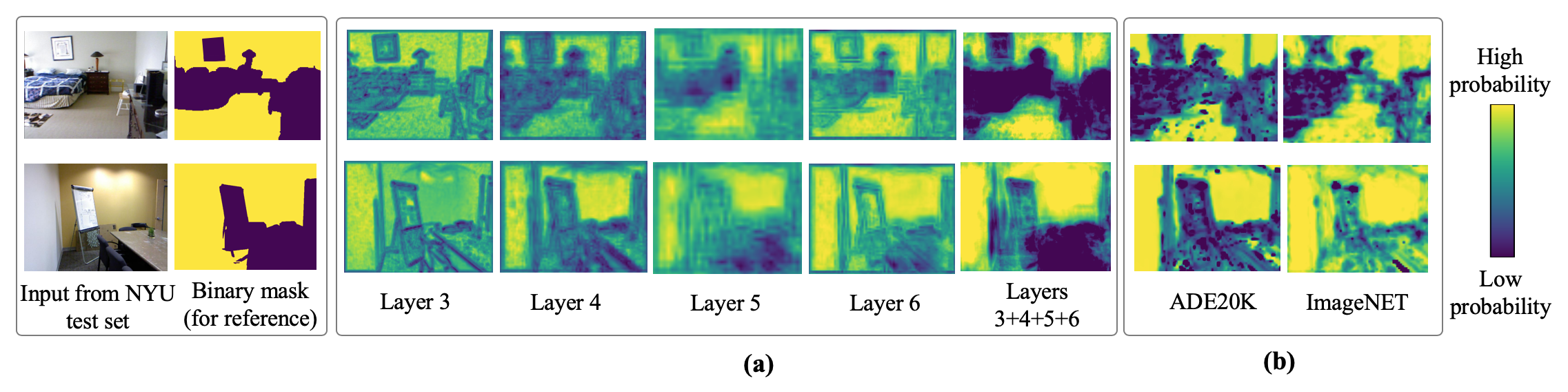}
    \vspace{-2mm}
    \caption{(a) Example of results on the NYU test set obtained by extracting features from the FCN at different layers (using the encoder trained on the NYU dataset). Results on deeper layers appear more blurry as the density has to be upsampled back to the input image size. The weighted average of layers 3 to 6 yields the best results, which is also reflected in Table \ref{tab:metricstestset}.
    Additionally (b) presents the results using ADE20K and ImageNet weights (at the fourth layer).}
    \label{fig:nyuresults}
\end{figure*}

\begin{table*}[t]
\caption{\label{tab:metricstestset} Evaluation on the NYU test set. Bold indicates the best performance for each approach and blue is the best overall.}
\centering
\scriptsize
\begin{tabular}{cc|ccc|ccc|ccc|ccc|}
\cline{3-14}
  &  & \multicolumn{3}{c|}{FPR@95\%TPR (\%)} & \multicolumn{3}{c|}{Average Recall (\%)} & \multicolumn{3}{c|}{Average Precision (\%)} & \multicolumn{3}{c|}{Area under ROC} \\ \cline{3-14} 
 &  & \begin{tabular}[c]{@{}c@{}}Processed \\ NYU\end{tabular} & \begin{tabular}[c]{@{}c@{}}ADE\\ 20K\end{tabular} & \begin{tabular}[c]{@{}c@{}}Image\\ Net\end{tabular} & \begin{tabular}[c]{@{}c@{}}Processed \\ NYU\end{tabular} & \begin{tabular}[c]{@{}c@{}}ADE\\ 20K\end{tabular} & \begin{tabular}[c]{@{}c@{}}Image\\ Net\end{tabular} & \begin{tabular}[c]{@{}c@{}}Processed \\ NYU\end{tabular} & \begin{tabular}[c]{@{}c@{}}ADE\\ 20K\end{tabular} & \begin{tabular}[c]{@{}c@{}}Image\\ Net\end{tabular} & \begin{tabular}[c]{@{}c@{}}Processed \\ NYU\end{tabular} & \begin{tabular}[c]{@{}c@{}}ADE\\ 20K\end{tabular} & \begin{tabular}[c]{@{}c@{}}Image\\ Net\end{tabular} \\ \hline
\begin{tabular}[ c]{@{}c@{}}FCN Softmax\end{tabular} & - & \textbf{45.20} & 54.9 & 52.3 & \textbf{37.4} & 37.0 & 37.0 & 81.7 & \textbf{83.0} & \textbf{83.0} & \textbf{0.87} & 0.86 & 0.86 \\ \hline
 & layer 3 & 69.5 & 79.3 & 86.5 & 27.3 & 16.2 & 13.1 & 75.3 & 64.2 & 61.1 & \textbf{0.79} & 0.69 & 0.65 \\
 & layer 4 & 67.3 & 66.2 & 71.5 & \textbf{28.6} & 22.0 & 19.8 & \textbf{76.1} & 70.1 & 67.8 & 0.77 & 0.75 & 0.73 \\
 & layer 5 & 71.2 & \textbf{64.6} & 71.6 & 24.6 & 21.7 & 19.8 & 72.6 & 70.0 & 67.8 & 0.76 & 0.76 & 0.73 \\
\multirow{-4}{*}{kNN} & layer 6 & 72.0 & 66.0 & 73.8 & 23.9 & 21.8 & 18.3 & 71.9 & 69.9 & 66.4 & 0.76 & 0.75 & 0.71 \\ \hline
\begin{tabular}[c]{@{}c@{}}kNN Ensemble\end{tabular} & 3+4+5+6 & 66.8 & \textbf{59.7} & 67.9 & \textbf{28.7} & 25.8 & 23.9 & \textbf{76.7} & 73.9 & 71.9 & \textbf{0.80} & 0.79 & 0.76 \\ \hline
 & layer 3 & 64.2 & 80.4 & 87.7 & 22.2 & 20.2 & 17.1 & 70.3 & 68.2 & 65.2 & 0.76 & 0.73 & 0.69 \\
 & layer 4 & 63.8 & 77.9 & 68.2 & 20.8 & 20.0 & 23.6 & 68.8 & 68.0 & 71.6 & 0.74 & 0.74 & 0.76 \\
 & layer 5 & 53.4 & 68.9 & 73.3 & 37.3 & 15.5 & 23.5 & 85.4 & 63.6 & 71.5 & \textbf{0.87} & 0.71 & 0.73 \\
\multirow{-4}{*}{\begin{tabular}[c]{@{}c@{}}Normalizing \\ Flow\end{tabular}} & layer 6 & \textbf{52.6} & 75.6 & 74.4 & \textbf{37.6} & 14.2 & 18.4 & \textbf{85.7} & 62.2 & 66.4 & 0.86 & 0.68 & 0.71 \\ \hline
Flow Ensemble & 3+4+5+6 & {\color{blue}\textbf{44.2}} & 70.9 & 70.6 & {\color{blue} \textbf{39.1}} & 16.0 & 24.7 & {\color{blue} \textbf{88.9}} & 65.7 & 74.4 & {\color{blue} \textbf{0.89}} & 0.72 & 0.76 \\ \hline
\end{tabular}
\end{table*}

\begin{table*}[t]
\caption{\label{tab:metrics} Evaluation on the generalization set. Bold indicates the best performance for each approach and blue is the best overall.}
\centering
\scriptsize
\begin{tabular}{cc|ccc|ccc|ccc|ccc|}
\cline{3-14}
 &  & \multicolumn{3}{c|}{FPR@95\%TPR (\%)} & \multicolumn{3}{c|}{Average Recall (\%)} & \multicolumn{3}{c|}{Average Precision (\%)} & \multicolumn{3}{c|}{Area under ROC} \\ \cline{3-14} 
 &  & \begin{tabular}[c]{@{}c@{}}Processed \\ NYU\end{tabular} & \begin{tabular}[c]{@{}c@{}}ADE\\ 20K\end{tabular} & \begin{tabular}[c]{@{}c@{}}Image\\ Net\end{tabular} & \begin{tabular}[c]{@{}c@{}}Processed \\ NYU\end{tabular} & \begin{tabular}[c]{@{}c@{}}ADE\\ 20K\end{tabular} & \begin{tabular}[c]{@{}c@{}}Image\\ Net\end{tabular} & \begin{tabular}[c]{@{}c@{}}Processed \\ NYU\end{tabular} & \begin{tabular}[c]{@{}c@{}}ADE\\ 20K\end{tabular} & \begin{tabular}[c]{@{}c@{}}Image\\ Net\end{tabular} & \begin{tabular}[c]{@{}c@{}}Processed \\ NYU\end{tabular} & \begin{tabular}[c]{@{}c@{}}ADE\\ 20K\end{tabular} & \begin{tabular}[c]{@{}c@{}}Image\\ Net\end{tabular} \\ \hline
\begin{tabular}[c]{@{}c@{}}FCN Softmax\end{tabular} & - & 80.6 & 74.7 & \textbf{65.0} & 27.0 & 41.7 & {\color{blue}\textbf{47.6}} & 40.8 & 60.5 & \textbf{63.4} & 0.77 & 0.81 & \textbf{0.86} \\ \hline
 \begin{tabular}[c]{@{}c@{}}kNN Ensemble\end{tabular} & 3+4+5+6 & 61.9 & \textbf{49.0} & 53.9 & 26.1 & 33.0 & \textbf{34.7} & 47.2 & 56.6 & \textbf{57.3} & 0.81 & \textbf{0.85} & 0.84 \\ \hline
 \begin{tabular}[c]{@{}c@{}}Flow Ensemble\end{tabular} & 3+4+5+6 & 56.4 & 49.4 & {\color{blue}\textbf{46.2}} & 38.0 & 27.0 & \textbf{40.3} & 61.6 & 51.8 & {\color{blue}\textbf{65.0}} & 0.84 & 0.83 & {\color{blue}\textbf{0.88}} \\ \hline
 \end{tabular}
\end{table*}

\subsection{Datasets}
We use the NYU Depth v2 dataset, which contains 1449 densely labeled images of indoor scenes. We sort the dataset to select images containing roughly as much background as foreground, retaining 580 images (493 for training, 87 for testing) with a balanced number of foreground and background pixels. We augment the data with flipping, rescaling, brightness and contrast changes. This dataset however only contains 464 different indoor scenes: a scene thus appears several times from different viewpoints, and could therefore be found in both our train and test set since they are randomly split. We therefore also created a dataset of 70 self-labeled indoor images, which also aims to better capture the high variability encountered in real life scenes. It includes: pictures taken with a smartphone, images downloaded online, black-and-white fish-eye images from construction sites, synthetic images, as well as collages (empty room images that we edited to add objects). It contains approximately 80\% background pixels. We use 45 images to create the \textit{generalization set} on which we will evaluate the performance. The 25 others are used as \textit{fitting dataset} to fit the logistic regression of the flow ensemble and kNN regresion.

\subsection{Evaluation Metrics}

Evringham et al. \cite{voc} suggest that measures such as accuracy are not well suited to evaluate classification and segmentation tasks as they are dependent on the prior distribution over the classes. In our binary segmentation problem, each pixel can be classified by comparing a score for background (coming from our FCN softmax or our density estimation) with a threshold. To choose this threshold, it has to be fit based on a prior assumption of the class-distribution, which is e.g. implicitly done with the cross-entropy loss on the training set. In an open-world setting, this prior distribution learned from the training data can be significantly different to the data used for evaluation. We therefore did not use precision or IoU to evaluate our method, but instead relied on Average Precision (AP) and Average Recall (AR) which allow for a fair comparison of the FCN softmax baseline and our density metric, evaluating over all possible thresholds and compensating for biases due to class distribution mismatches. AR score shows our ability to detect foreground objects, while the AP score shows the overall segmentation quality. The popular VOC Challenge \cite{voc} for example uses AP as their ranking metric for detection, classification, and segmentation. 

We also report the Area Under the Receiver Operating Characteristic Curve (AUROC) and the False Positive Rate (FPR) at 95\% True Positive Rate (TPR), both standard metrics for outlier detection \cite{metrics, ood2, distancebased}. The FPR at 95\% TPR is a particular informative metric for safety-critical applications. The TPR (or recall) is the fraction of foreground pixels correctly labeled. For safety reasons, we value having a low number of misclassified foreground pixels. We thus ensure having a high TPR (95\%) and then report the corresponding FPR value. The AUROC can be interpreted as the probability that a positive example has a greater detector score/value than a negative example. A random segmentation thus has an AUROC of $0.5$ while a perfect segmentation will equal $1$ \cite{metrics}. 

\subsection{Results on the NYU Test Set}
\label{subsec:results_test} 
In this section, we have a closer look at how our results on the NYU test set are influenced by two main factors: the layers of the FCN from which the background features are extracted and the weights used for the FCN's encoder (NYU segmentation, ImageNet or ADE20K). We qualitatively show these results in Fig.~\ref{fig:nyuresults}, and quantitatively in Table~\ref{tab:metricstestset}.

Using the encoder trained on the NYU training dataset yields, as expected, the best results on the NYU test set. When comparing ImageNet and ADE20K, the former is slightly superior for normalizing flow while the latter has a small advantage for kNN. Overall, ImageNet and ADE20K perform similarly on the NYU test set.

Table~\ref{tab:metricstestset} clearly shows a superior performance of normalizing flow with respect to kNN. It also suggest that by using deeper layers in the encoder, the features become more distinctive and objects can be better identified. Finally, using a combination of all the layers further improves the results and makes normalizing flows superior to the classical softmax segmentation on all evaluated metrics. Without ever seeing the objects in the NYU dataset during training, our method surpasses the softmax segmentation (which included the objects in training).

The important point from the experiment on the NYU test set is that our method does not come with a performance trade-off when the testing data is similar to the training (it is even better). The superiority of our method is further demonstrated using the novel generalization set (see table \ref{tab:metrics}, section \ref{subsec:generalization}). 

\subsection{Results on the Generalization Set}
\label{subsec:generalization}
\begin{figure}[t]
	\centering
    \includegraphics[width=0.70\columnwidth]{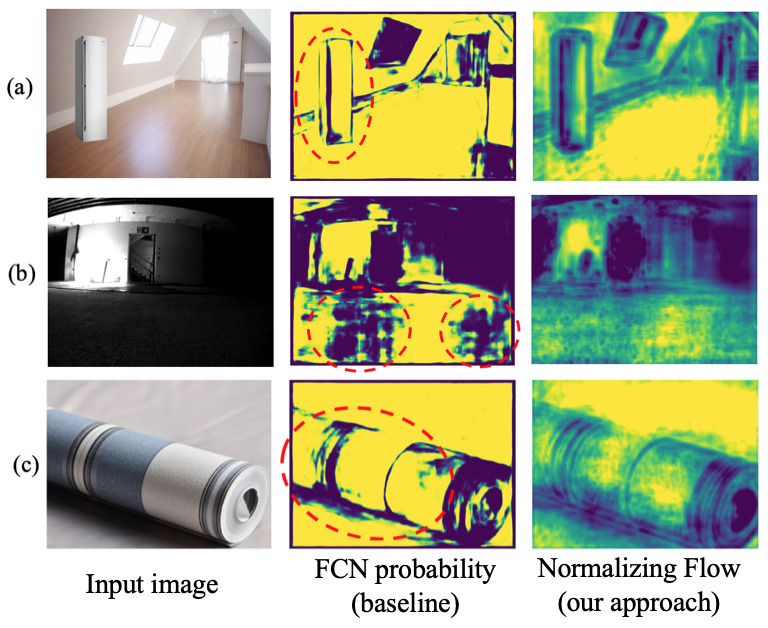}
	\caption{Example results on the generalization set (red dotted circles were added to highlight FCN softmax failures). (a) Empty room image from Google where we added a fridge: it was not found by the FCN softmax. (b) Greyscale fish-eye image from a construction site. The NYU dataset contains neither greyscale nor fish-eye images, which explains the FCN softmax's failure. Despite the difficulty, the density estimation works remarkably well. (c) Advertisement image where part of the textile roll is missed by the FCN softmax. This is challenging since the NYU dataset does not contain close views of objects. Note that these images are more difficult than the ones in figure \ref{fig:method_test}, which explains the slightly poorer results.}
	\label{fig:generalization}
\end{figure}

We use the generalization set to show how well our method generalizes to novel objects. Fig. \ref{fig:generalization} shows a few examples from this set where the FCN fails to segment objects outside of the training set, while our method does not suffer from this. Table \ref{tab:metrics} quantitatively shows that our method outperforms FCN softmax and kNN on the generalization set. 

As expected, when using an encoder trained on the small NYU dataset the softmax segmentation has extremely poor results on novel data. Using normalizing flow with features from the same encoder however yields far better segmentation results. Those can be seen on Table~\ref{tab:metrics}, the average recall is improved from 27\% to 38\% and the average precision is improved from 41\% to 62\%. This is the key element of our contribution, showing the benefit of learning the background feature distribution using normalizing flow. 

Additionally, we notice that the encoder trained on ImageNet unquestionably yields the best generalization results. While intuitively one could think that features from ADE20K would be more relevant for our task since ADE20K is a segmentation dataset, our experiments show the opposite. Although ADE20K contains approx. 20K images, it does not capture the same variability of scenes as ImageNet (14 millions images). This explains the superior behaviour of ImageNet weights on our more diverse generalization dataset. Therefore that size and diversity of the encoder training set directly influence the generalization results


\subsection{Computational Aspects}
\label{subsec:computation}

Normalizing flow also has the advantage of being computationally less expensive than kNN. Table~\ref{tab:computation} shows results using an NVIDIA GeForce GTX 1080 Ti GPU. It is important to note that neither normalizing flow nor kNN were optimized for fast computation. Our experiments show that using 16 bijectors instead of 32 significantly would reduce the flow computation, with minimal effect on segmentation quality. Moreover kNN could be optimized to make better use of GPU resources~\cite{kNN_gpu}.

For kNN, we stored a different amount of features depending on the layer, which explains the differences in time and memory. Even with the smallest amount of stored features (layer 5), kNN requires more than 5X more time and memory.

The time needed by our flow based method decreases as we use deeper layers, since the feature map is reduced after each pooling layer. Note that the $6^{th}$ layer has the same feature map size as the $4^{th}$ layer but the dimension is reduced from 512 to 128. Using a normalizing flow at this layer is thus faster and cheaper in memory than using the $4^{th}$ layer. It is even faster than using the FCN decoder. 

\begin{table}
\caption{\label{tab:computation} Evaluation off Inference Time and Model Size}
\centering
\scriptsize
\resizebox{\columnwidth}{!}{%
\begin{tabular}{|c|c|c|c|}
\hline
& \textbf{FCN Softmax} 
& \textbf{Flow Density} 
& \textbf{kNN Density} \\ 
\hline
Time [ms] & 35 & 100-50-30-40  & 1950-770-250-370 \\ \hline
Memory [MB] & 160 & 160-230-230-130 * & 11000-5100-1360-2000 \\ \hline
\multicolumn{4}{l}{} \\
\multicolumn{4}{l}{\textit{Note}: Flow and kNN density results are presented as layers 3-4-5-6} \\
\multicolumn{4}{l}{*These corresponds to the memory of the flow ensemble weights (without the encoder).} 
\end{tabular}%
}
\end{table}


\section{Discussion}

Using kNN as a density estimation method has given poor results (see tables \ref{tab:metricstestset} and \ref{tab:metrics}) since it is a poor high-dimensional density estimation method compared to normalizing flow. kNN can indeed not model the complex high dimensional probability distribution of background features so we will focus the rest of this discussion on our flow-based approach.

 We have successfully proven that, given the same encoder features, our normalizing flow approach is superior to a classical deconvolution with softmax. To further improve the segmentation, more advanced experts network could be used (e.g. DeepLab) since we expect the normalizing flow method to improve as features from the expert network also improve. 

Our experiments with ImageNet and ADE20K weights revealed that specialized training of the expert network is not necessary. However, we observed training instabilities of the normalizing flow when using ImageNet or ADE20K weights. The main difference of those to the NYU weights is that they do not use batch normalization. This spreads out features in the high-dimensional space and makes training of the density estimation much harder. To train the flow, we therefore used early stopping when the log-likelihood of the NYU validation started to decrease. That limited the modeling capacities of the normalizing flow and can explain the relatively poor results of ImageNet and ADE20K on the training set (Table \ref{tab:metricstestset}). 

\addtolength{\textheight}{-1.5cm}   

Deeper layers of the encoder have a smaller feature size which allows faster inference. Table~\ref{tab:computation} shows that flow-based segmentation on the features of the $5^{th}$ layer is even faster than using the FCN decoder presented in Figure~\ref{fig:method_a}. Using an embedding layer for dimensionality reduction would further reduces the runtime and memory of our flow approach and would make training of normlizing flow more stable. However, since dimensionality reduction compresses the information, extensive experiments are needed to assess whether using an embedding layer before the normalizing flow would make the separation of classes in the lower dimensional space harder, leading to poorer segmentation results. We encourage future works to evaluate the benefits of using embedding layers.  

A limitation of our method is that similarly to the FCN softmax, planar and texture-less surfaces are often mistakenly labeled as background with high certainty. Another limitation is that background not found in the training set will be labeled as foreground. For example, some types of floor (e.g. parquet) are often thought to be less likely to be background than walls. This limitation however emphasizes a strength of our method: given data that has no support in the training set, we recognise it as unlikely. As we motivated our work with the detection of potentially dynamic objects for autonomous robots, this is a desirable behavior. While a conventional segmentation network produces random classifications for unknown objects, density based segmentation assigns a low likelihood given the support of known static elements. Any mapping or planning algorithm using this information therefore will not assume unknown structure to be static. Labelling OoD background as (potential) foreground may limit segmentation quality but improves safety and reliability for the robotic system.

To further improve the performance on OoD background, we suggest using data augmentation. Since background varies in textures rather than shapes, the same images with changing textures of the walls/floors/ceiling etc. can be used. Such augmentation can be simulated, thus reducing the need for more annotated data.

Conventional segmentation methods have weaknesses on both OoD foreground and background since they behave randomly when encountering unknown data. Density based segmentation successfully finds OoD foreground and it's failure on OoD background is predictable for downstream applications. These are important improvements over classical methods towards safer and more reliable autonomous mobile robots.

\section{Conclusion}
\label{sec:conclusion}
We presented a novel approach to segment foreground from background in indoor environments, with high reliability against the variability of indoor scenes. Unlike any existing works, we learn the distribution of background features with flow-based methods in order to distinguish foreground objects. The combination of several layers further increases the performance of our approach and yields the best results. We demonstrated that our method outperforms classical softmax-based segmentation on diverse and novel images without any performance trade-offs on the NYU test set. This is a critical advantage for the deployment in real life applications.

Our reliable segmentation of static background has potential impact on a variety of robotic tasks, e.g. novelty detection in indoor scenes, obstacle avoidance by recognizing arbitrary objects, or long-term SLAM localization by reliably segmenting static structure from dynamic objects. We see all of these as exciting directions for future research.

\end{document}